\documentclass[runningheads]{llncs}
\usepackage{graphicx}
\usepackage{subfigure}
\usepackage{comment}
\usepackage{url}
\usepackage{wrapfig,lipsum,booktabs}
\usepackage{amsmath,amssymb}
\begin{document}
%
\title{Semi-Supervised Models via Data Augmentation for Classifying Interactive Affective Responses}
\titlerunning{Semi-Supervised Models via Data Augmentation}
%
\author{Jiaao Chen $^*$\inst{1} \and
Yuwei Wu\thanks{Equal Contribution. This work was done when Yuwei is a visiting student at Georgia Tech.}\inst{2} \and
Diyi Yang\inst{1}}
\authorrunning{C. Jiaao and W. Yuwei et al.}
%
\institute{Georgia Institute of Technology, Atlanta GA 30318, USA \and
Shanghai Jiao Tong University, Shanghai, China\\
\email{jiaaochen@gatech.edu, will8821@sjtu.edu.cn, diyi.yang@cc.gatech.edu}}
\maketitle              
\begin{abstract}
We present semi-supervised models with data augmentation (SMDA), a semi-supervised text classification system to classify interactive affective responses. SMDA utilizes recent transformer-based models to encode each sentence and employs back translation techniques to paraphrase given sentences as augmented data. For labeled sentences, we performed data augmentations to uniform the label distributions and computed supervised loss during training process. For unlabeled sentences, we explored self-training by regarding low-entropy predictions over unlabeled sentences as pseudo labels, assuming  high-confidence predictions as labeled data for training. We further introduced consistency regularization as unsupervised loss after data augmentations on unlabeled data, based on the assumption that the model should predict similar class distributions with original unlabeled sentences as input and augmented sentences as input. Via a set of experiments, we demonstrated that our system outperformed baseline models in terms of F1-score and accuracy.
\end{abstract}

\section{Introduction}
\textbf{Affect} refers to emotion, sentiment, mood, and attitudes including subjective evaluations, opinions, and speculations \cite{articleaffective}.
Psychological models of affect have been utilized by other extensive computational research to operationalize and measure users' opinions, intentions, and expressions. 
Understanding affective responses with in conversations is an important first step for studying affect and has attracted a growing amount of research attention recently \cite{yang2019channel,ernala2017linguistic,yang2019seekers}. The affective understanding of conversations focuses on the problem of how speakers use emotions to react to a situation and to each other, which can help better understand human behaviors and build better human-computer-interaction systems.

However, modeling affective responses within conversations is relatively challenging since it is hard to quantify the affectiveness \cite{pub.1084126691} and there are no large-scale labeled dataset about affective levels in responses. In order to facilitate research in modeling interactive affective responses, \cite{overview_claff2020} introduced a conversation dataset, OffMyChest, building from Reddit, and proposed two tasks: (1) Semi-supervised learning task: predict labels for Disclosure and Supportiveness in sentences based on a small amount of labeled and large unlabeled training data; (2) Unsupervised task: design new characterizations and insights to model conversation dynamics. The current work focused on the first task. 

With limited labeled data and large amount of unlabeled data being given, to alleviate the dependence on labeled data, we combine recent advances in language modeling,  semi-supervised learning on text and data augmentations on text to form Semi-Supervised Models via Data Augmentation (SMDA). SMDA consists of two parts: supervised learning over labeled data (Section~\ref{sup}) and unsupervised learning over unlabeled data (Section~\ref{unsup}). Both parts utilize data augmentations to enhance the learning procedures. 
Our contributions in this work can be summarized into three parts: analysed the OffMyChest dataset in Section~\ref{anaysis}, proposed a semi-supervised text classification system to classify interactive affective responses classification in Section~\ref{reference} and described the experimental details and results in Section~\ref{result}. 
 
\section{Related Work}

\paragraph{Transformer-based Models}:
With transformer-based pre-trained models becoming more and more widely-used, pre-training and fine-tuning framework \cite{howard-ruder-2018-universal} with large pre-trained language models are applied into a lot of NLP applications and achieved state-of-the-art performances \cite{xie2019unsupervised}. Language models \cite{peters-etal-2018-deep,howard-ruder-2018-universal,abs-1906-08237}  or masked language models \cite{devlin-etal-2019-bert,abs-1901-07291} are pre-trained   over a large amount  of  text  from  Wikipedia and then fine-tuned on specific tasks like text classifications. Here we built our SMDA system based on such framework.

\paragraph{Data Augmentation on Text}:
When the amount of labeled data is limited, one common technique for handling the shortage of data is to augment given data and generate more training ``augmented'' data. Previous work has utilized simple operations like synonym replacement, random insertion, random swap and random deletion for text data augmentation  \cite{abs-1901-11196}. Another line of research applied neural models for augmenting text by generating paraphrases via back translations \cite{xie2019unsupervised} and monotone submodular function maximization \cite{kumar-etal-2019-submodular}. Building on those prior work, we utilized back translations as our augment methods on both labeled and unlabeled sentences.

\paragraph{Semi-Supervised Learning on Text Classification}: 
One alternative to deal with the lack of labeled data is to utilize unlabeled data in the learning process, which is denoted as Semi-Supervised Learning (SSL), since unlabeled data is usually easier to get compared to labeled data. Researchers has made use of variational auto encoders (VAEs)  \cite{mchen-variational-18,YangHSB17,abs-1906-02242}, self-training  \cite{article:Pseudo-Label,Grandvalet:2004:SLE:2976040.2976107,Meng:2018:WNT:3269206.3271737}, consistency regularization \cite{MiyatoMKI19,shen2018deep,xie2019unsupervised} to introduce extra loss functions over unlabeled data to help the learning of labeled samples. VAEs utilize latent variables to reconstruct input labeled and unlabeled sentences and predict sentence labels with these latent variables; self-training adds unlabeled data with high-confidence predictions as pseudo labeled data during training process and consistency regularization forces model to output consistent predictions after adding adversarial noise or performing data augmentations to input data. We combined self-training, entropy minimization and consistency regularization in our system for unlabeled sentences.

\section{Data Analysis and Pre-processing} \label{anaysis}
Researching how human initiate and hold conversations has attracted increasing attention those days, as it can help us better understand how human behave over conversations and build better AI systems like social chatbot to communicate with people.  
In this section, we took a closer look at the conversation dataset, OffMyChest \cite{overview_claff2020}, for better understanding and modeling interactive affective responses. Specifically, we describe certain characteristics of this dataset and our pre-processing steps.

\subsection{Label Definition}
For each comment of a post on Reddit, \cite{overview_claff2020} annotated them with 6 labels: \textit{Information\_disclosure} representing some degree of personal information in comments; \textit{Emotional\_disclosure} representing comments containing certain positive or negative emotions; \textit{Support} referring to comments offering social support like advice; \textit{General\_support} representing that comments are offering general support through quotes and catch phrases, with \textit{Information\_support} offering specific information like practical advice, and  \textit{Emotional\_support} offering sympathy, caring or encouragement. Each comment can belong to multiple categories. 

\begin{figure}
\includegraphics[width=\textwidth]{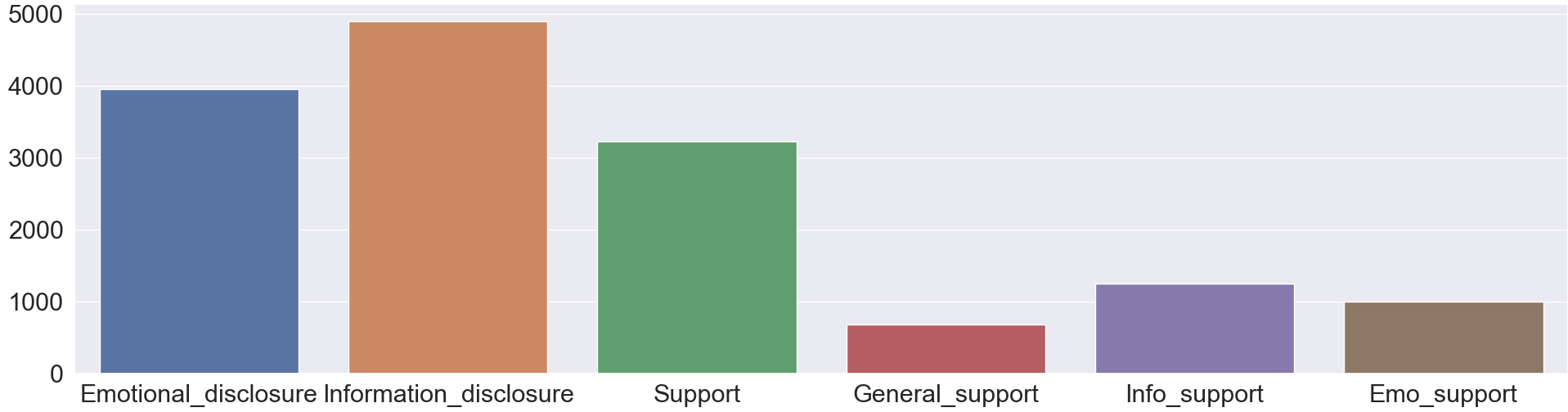}
\caption{Distribution of each label in the labeled corpus. The $y$ axis is the number of sentences that have the corresponding labels.} \label{fig:dist}
\end{figure}

\begin{table}
\centering
 \caption{Dataset split statistics. We utilized both labeled data and unlabeled data for training, generated dev and test set by sampling from given labeled comments set.}\label{tab:split}
\begin{tabular}{c|c|c|c}
\hline
\textbf{Labeled Train Set} &  \textbf{Dev Set}& \textbf{Test Set} & \textbf{Unlabeled Train Set}\\
\hline \hline
 {8,000} &  {2,000} & {2,860}& {420,607}\\
\hline
\end{tabular}
\end{table}

\newcommand{\tabincell}[2]{\begin{tabular}{@{}#1@{}}#2\end{tabular}}
\begin{table}
\centering
 \caption{Paraphrase examples generated via back translation from original sentences into augmented sentences.}\label{tab:exp}
\begin{tabular}{l|l||l}
\hline
\textbf{Original} &  \textbf{Augmented} & \textbf{Labels}\\
\hline \hline
 \tabincell{l}{I’m crying a lot of tears of joy \\right now.}&  \tabincell{l}{Right now I’m crying a lot of \\ happy tears.} &\tabincell{l}{Emo\_disclosure}\\
\hline
 \tabincell{l}{Stepdad will be the one walking \\ me down the aisle when I get\\ married.}&  \tabincell{l}{It will be my stepfather walking \\me down the aisle when I\\ get married.} &\tabincell{l}{Info\_disclosure}\\
\hline
 \tabincell{l}{Hope you have a nice day.}&  \tabincell{l}{I hope you have a good day.} &\tabincell{l}{Support}\\
\hline
 \tabincell{l}{Your best effort, both of you}&  \tabincell{l}{Both of you are giving it your \\best shot.} &\tabincell{l}{General\_support}\\
\hline
 \tabincell{l}{Plan your transition back to \\working outside of the home.}&  \tabincell{l}{Plan your move back to a job \\outside your own home.} &\tabincell{l}{Info\_support}\\
\hline
\tabincell{l}{I am so freaking happy for you!}&  \tabincell{l}{ I’m so excited for you!} &\tabincell{l}{Emo\_support}\\
\hline
\end{tabular}
\end{table}

\subsection{Data Statics}
In OffMyChest corpus, there are 12,860 labeled sentences and over 420k 
unlabeled sentences for training, 5,000 unlabeled sentences for test. The 
label distributions of labeled sentences are showed in Fig.~\ref{fig:dist}.
To train and evaluate our systems, we randomly split the given labeled sentence set into train, development and test set. The data statics are shown in Table~\ref{tab:split}. We tuned hyper-parameters and chose best models based on performance on dev set, and reported model's performance on test set. 
\begin{figure}
\centering
\includegraphics[width=0.5 \textwidth]{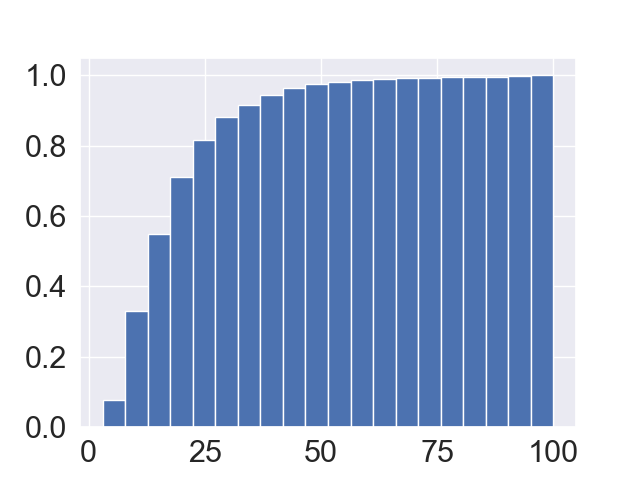}
\caption{Cumulative distribution of sentence length in the given labeled sentence set. The $y$ axis represents the portion over all sentences.} \label{fig:length}
\end{figure}
\subsection{Pre-processing}
We utilized XLNet-cased-based tokenizer\footnote{\url{https://huggingface.co/transformers/model_doc/xlnet.html#xlnettokenizer}} to split each sentence into tokens. We showed the cumulative sentence length distribution in Fig.~\ref{fig:length}, 
95\% comments have less than 64 tokens. Thus we set the maximum sentence length to 64, and remained the first 64 tokens for sentences that exceed the limit. As for data augmentations, we made use of back translation with German as middle language to generate paraphrases for given sentences. Specifically, we loaded translation model from Fairseq\footnote{\url{https://github.com/pytorch/fairseq}}, translated given sentences from English to German, and then translated them back to English. Also to increase the diversity of generated paraphrases, we employed random sampling with a tunable temperature (0.8) instead of beam search for the generation. We describe some examples in Table~\ref{tab:exp}.

\section{Method} \label{reference}
We convert this 6-class affective response classification task into 6 binary classification tasks, namely whether each sentence belongs to each category or not (labeled with 1 or 0). For each binary classification task, given a set of labeled sentences consisting of $n$ samples $S = \{s_1, ..., s_n\}$ with labels $L = \{l_1, ..., l_n\}$, where $l_i \in \{0,1\}^2$, and a set of unlabeled sentences $S_u = \{s_1^u, ..., s_m^u\}$, our goal is to learn the classifier $f(\hat{l}|s,\theta_i), i \in [1,6]$.
Our SMDA model contains several components: Supervised Learning (Section~\ref{sup}) for labeled sentences, Unsupervised Learning (Section~\ref{unsup}) for unlabeled sentences, and Semi-Supervised Objective Function (Section~\ref{ssl}) to combine labeled and unlabeled sentences.

\begin{figure*}[t!]
\centering
\subfigure[.3\linewidth][Before Augmentation]{%
\label{fig:before}
\includegraphics[width=0.88\textwidth]{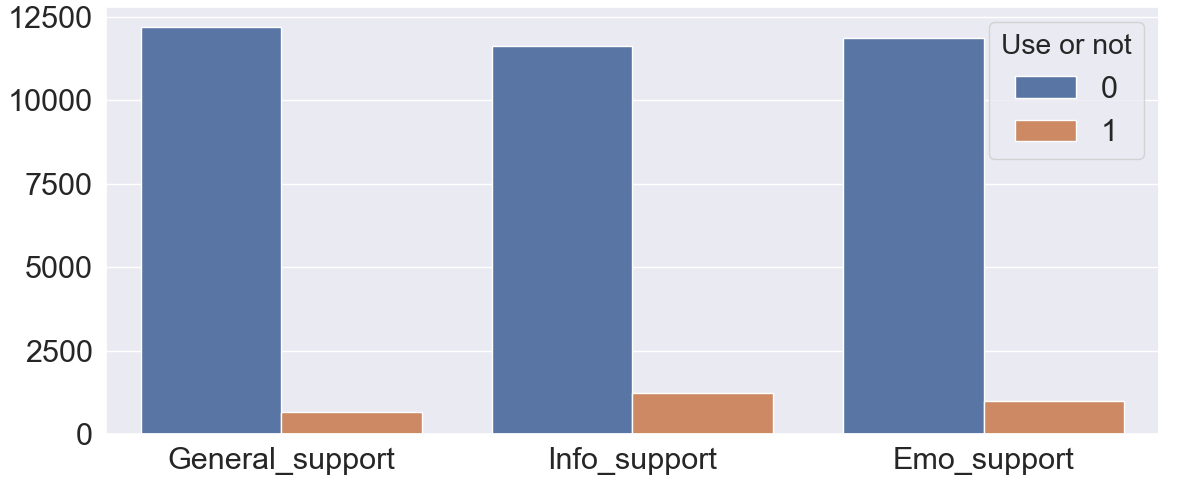}} 
\subfigure[.48\linewidth][After Augmentation]{\label{fig:after}

\includegraphics[width=0.88\textwidth]{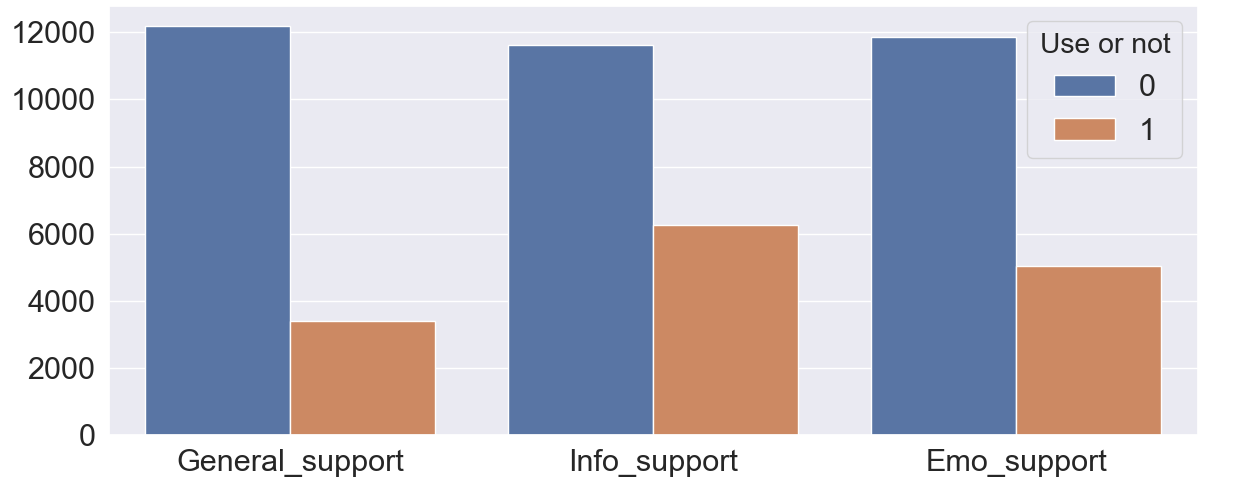}}
\caption[]{Distributions before and after performing augmentations over labeled sentences belonging to \textit{General\_support}, \textit{Info\_support} and \textit{Emo\_support}. 0 means sentences don't use corresponding types of support, while 1 represents sentences use corresponding types of support. $y$ axis is the number of sentences.}
\label{fig:dist_compare}
\end{figure*}

\subsection{Supervised Learning} \label{sup}
\subsubsection{Generating Balanced Labeled Training Set}
As shown in Fig.~\ref{fig:dist}, the distribution is very unbalanced with respect to \textit{General\_support}, \textit{Info\_support} and \textit{Emo\_support}. In order to get more training sentences with these three types of support and make these three binary classification sub-tasks learn-able with a more balanced training set, we performed data augmentations over sentences with these three labels. Specifically, we paraphrased each sentence by 4 times via back translations and regarded that the augmented sentences have the same labels as original sentences. The comparison distributions are shown in Fig.~\ref{fig:dist_compare}

\subsubsection{Supervised Learning for Labeled Sentences}
For each input labeled sentence $s_i$,  we used XLNet \cite{abs-1906-08237} $g(.)$ to encode it into hidden representation $h_i = g(s_i)$, and then passed them though a 2-layer MLP to predict the class distribution $\hat{l_i} = f(h_i)$. Since these sentences have specific labels, we optimize the cross entropy loss as supervised loss term:
\begin{equation}
    L_S(s_i, l_i) = - \sum l_i \log f(g(s_i))
\end{equation}

\subsection{Unsupervised Learning} \label{unsup}

\subsubsection{Paraphrasing Unlabeled Sentences} We first performed back translations once for each unlabeled sentence $s_i^u \in S_u$ to generate the augmented sentence set $S_{u,a} = \{s_1^{u,a}, ..., s_m^{u,a}\}$ in the same manner we described before.

\subsubsection{Guessing Labels for Unlabeled Sentences} For an unlabeled sentence $s_i^u$, we utilized $g(.)$ and $f(.)$ in Section~\ref{sup} to predict the class distribution:
\begin{equation}
    \hat{l}_i^u = f(g(s_i^u))
\end{equation}
To avoid the prediction being so close to uniform distribution, we generate low-entropy guessing labels $\tilde{l}_i^u$ by a sharpening function \cite{abs-1905-02249}:
\begin{equation}
    \tilde{l}_i^u = \frac{(\hat{l}_i^u)^{\frac{1}{T}}}{||(\hat{l}_i^u)^{\frac{1}{T}}||_1}
\end{equation}
where $||.||_1$ is $l_1$-norm of the vector. When $T\to 0$, the guessed
label becomes an one-hot vector. 

\subsubsection{Self-training for Original Sentences} Inspired by self-training where model is also trained over unlabeled data with high-confidence predictions as their labels, in SMDA, with our guessed labels $\tilde{l}_i^u$ with respect to original unlabeled sentence $s_i^u$, we added such pair $(s_i^u, \tilde{l}_i^u)$ into training by minimize the KL Divergence between them:
\begin{equation}
    L_s(s_i^u) = KL (f(g(s_i^u)) ||  \tilde{l}_i^u )
\end{equation}

\subsubsection{Entropy Minimization for Original Sentences} One common assumption in many semi-supervised learning methods is that a classifier's decision boundary should not pass through high-density regions of the marginal data distribution \cite{Grandvalet:2004:SLE:2976040.2976107}. Thus for original unlabeled sentence $s_i^u$, we added another loss term to minimize the entropy of model's output:
\begin{equation}
    L_e(s_i^u) = - \sum f(g(s_i^u)) \log f(g(s_i^u))
\end{equation}

\subsubsection{Consistency Regularization for Augmented Sentences} With the assumption that the model should predict similar distributions with input sentences before and after augmentations, we minimized the KL Divergence between outputs with original sentence $s_i^u$ as input and augmented sentence $s_i^{u,a}$ as input:
\begin{equation}
    L_c(s_i^u) =  KL (f(g(s_i^{u,a})) || \hat{l}_i^u  )
\end{equation}

\noindent
Combining all the loss terms for unlabeled sentences, we defined our unsupervised loss terms as:
\begin{equation}
    L_U(s_i^u) = L_s(s_i^u)  + L_e(s_i^u) + L_c(s_i^u)
\end{equation}

\subsection{Semi-Supervised Objective Function} \label{ssl}
We combined the supervised and unsupervised learning described above to form our overall semi-supervised objective function:
\begin{equation}
    L = \mathbb{E}_{(s_i, l_i) \in (S,L) } L_S(s_i, l_i) + \gamma \mathbb{E}_{s_i^u \in S_u} L_U(s_i^u)
\end{equation}
where $\gamma$ is the balanced weight between supervised and unsupervised loss term.

\section{Experiments}\label{result}
\subsection{Model Setup}
In SMDA \footnote{The codes and data split will be released later.}, we only used single model for each task without jointly training and parameter sharing. That is, we trained six separate classifiers on these tasks. Inspired by recent success in pre-trained language models, we utilized the pre-trained weights of XLNet and followed the same fine-tuning procedure as XLNet. We set the initial learning rate for XLNet encoder as 1e-5 and other linear layers as 1e-3. The batch size was selected in $\{32, 64, 128, 256\}$. The maximum number of epochs is set as 20. Hyper-parameters were selected using the performance on development set. The sharpen temperature $T$ was selected in $\{0.3, 0.5, 0.8\}$ depending on different tasks. The balanced weight $\gamma$ between supervised learning loss and unsupervised learning loss term started from a small number and grew through training process to 1.

\subsection{Results}
Our experimental results are shown in Table~\ref{tab:result}. We compared our proposed SMDA with BERT and XLNet in terms of accuracy(\%) and Macro F1 score. BERT and XLNet achieved similar performance since they both obey the pre-training and fine-tuning manner. When combining with augmented and more balanced labeled data, massive unlabeled data, our SMDA achieved best performance across six binary-classification tasks. And we submitted the classification results on given unlabeled test set.

\begin{table*}
	\centering 
		\caption{\label{tab:result} Results on test set. Our baseline is our implementation of XLNet-cased-base.
	}
	\resizebox{\linewidth}{!}
	{
		\begin{tabular}{l|cccccccccccc}
			\hline
			
			\hline
			\textbf{Task} &  \multicolumn{2}{c}{\textbf{Emo\_disc}} &\multicolumn{2}{c}{\textbf{Info\_disc}\text{ }} & \multicolumn{2}{c}{\textbf{Support}\text{ }} &  \multicolumn{2}{c}{\textbf{Gen\_supp}\text{ }}&\multicolumn{2}{c}{\textbf{Info\_supp}\text{ }}&\multicolumn{2}{c}{\textbf{Emo\_supp}\text{ }}\\
			&acc & F1	& acc & F1 & acc & F1& acc & F1& acc & F1& acc & F1\\
         \hline

            BERT$_\text{BASE}$ & 71.3 & 65.7 & 71.1 & 68.7 & 81.9  & 75.6 & 90.6 & 63.9 &  88.9&69.8&92.9&73.8\\
			XLNet$_\text{BASE}$   & 72.4 & 67.9 & 72.2 & 69.3 & 83.4  & 77.3 & \textbf{92.7} & \textbf{65.0} &  87.9&70.3&93.4&73.8\\
			SMDA & \textbf{75.2} & \textbf{68.5} &\textbf{ 74.3} & \textbf{71.0} & \textbf{83.5} & \textbf{77.7} & 91.7 & 63.7 & \textbf{89.9} & \textbf{70.5} & \textbf{93.6} & \textbf{76.2}\\
			\hline
			
			\hline
		\end{tabular}
	}

\end{table*}
\section{Conclusion}
In this work, we focused on identifying disclosure and supportiveness in conversation responses based on a small labeled and large unlabeled training data via our proposed semi-supervised text classification system : Semi-Supervised  Models  via  Data  Augmentation (SMDA). SMDA utilized supervised learning over labeled data and conducted self-training, entropy minimization and consistency regularization over unlabeled data. Experimental results demonstrated that our system outperformed baseline models significantly. 
%
%
%
\bibliographystyle{splncs04}
\bibliography{cite}

\end{document}